\documentclass[10pt,twocolumn,letterpaper]{article}

\usepackage{wacv}              

\usepackage{times}    
\usepackage{xspace}
\usepackage[dvipsnames]{xcolor}
\usepackage{graphicx}
\usepackage{amsmath}
\usepackage{amssymb}
\usepackage{booktabs}
\usepackage{multirow}
\usepackage{float}
\usepackage[ruled,vlined]{algorithm2e}
\usepackage[numbers,sort&compress]{natbib}

\usepackage[pagebackref,breaklinks,colorlinks]{hyperref}

\usepackage[capitalize]{cleveref}
\crefname{section}{Sec.}{Secs.}
\Crefname{section}{Section}{Sections}
\Crefname{table}{Table}{Tables}
\crefname{table}{Tab.}{Tabs.}

\def\wacvPaperID{2760} 
\def\confName{WACV}
\def\confYear{2025}

\begin{document}


\title{From Pixels to Privacy: Temporally Consistent Video Anonymization via Token Pruning for Privacy Preserving Action Recognition}

\author{
Nazia Aslam$^{1}$ \hspace{0.2cm}
Abhisek Ray$^{2}$ \hspace{0.2cm}
Joakim Bruslund Haurum$^{3,4}$ \hspace{0.2cm}
Lukas Esterle$^{2}$ \hspace{0.2cm}
Kamal Nasrollahi$^{1,5}$ \vspace{0.2cm}
\\
$^{1}$Aalborg University, Denmark \hspace{0.2cm}
$^{2}$Aarhus University, Denmark \hspace{0.2cm}
$^{3}$Poineer Centre for AI, Denmark \\
$^{4}$University of Southern Denmark, Denmark \hspace{0.2cm}
$^{5}$Milestone System, Denmark \\
{\tt\small \{naas, kn\}@create.aau.dk, \{ar, lukas.esterle\}@ece.au.dk, jhau@mmmi.sdu.dk}
}

\maketitle

\begin{abstract}

Recent advances in large-scale video models have significantly improved video understanding across domains such as surveillance, healthcare, and entertainment. However, these models also amplify privacy risks by encoding sensitive attributes, including facial identity, race, and gender. While image anonymization has been extensively studied, video anonymization remains relatively underexplored, even though modern video models can leverage spatiotemporal motion patterns as biometric identifiers. To address this challenge, we propose a novel attention-driven spatiotemporal video anonymization framework based on systematic disentanglement of utility and privacy features. Our key insight is that attention mechanisms in Vision Transformers (ViTs) can be explicitly structured to separate action-relevant information from privacy-sensitive content. Building on this insight, we introduce two task-specific classification tokens, an action CLS token and a privacy CLS token, that learn complementary representations within a shared Transformer backbone. We contrast their attention distributions to compute a utility–privacy score for each spatiotemporal tubelet, and keep the top-$k$ tubelets with the highest scores. This selectively prunes tubelets dominated by privacy cues while preserving those most critical for action recognition. Extensive experiments demonstrate that our approach maintains action recognition performance comparable to models trained on raw videos, while substantially reducing privacy leakage. These results indicate that attention-driven spatiotemporal pruning offers an effective and principled solution for privacy-preserving video analytics. Find code \href{https://github.com/Rabusi/From-Pixels-to-Privacy-Temporally-Consistent-Video-Anonymization-via-Token-Pruning-for-Privacy-Pres/tree/main}{HERE}.
     
\end{abstract}


\section{Introduction}
\label{sec:intro}

Video understanding is a fundamental research area in computer vision, covering a wide range of tasks, including human action recognition, anomaly detection, and temporal action localization. The ability to effectively model the spatiotemporal cues is essential for numerous real‑world applications, including patient monitoring, sports analytics, robotics, and intelligent surveillance systems. Recent models have increasingly focused on learning transferable spatiotemporal representations that generalize across various video understanding tasks \cite{wang2023videomae, wang2022internvideo, bardes2023v}. These models are trained on large-scale video datasets to capture the appearance and motion patterns. However, these models raise privacy concerns because they can implicitly encode sensitive personal attributes, such as facial features, gender, and race, within their learned representations. The large-scale collection and automated analysis of such data have motivated stringent regulatory frameworks, including the General Data Protection Regulation (GDPR) \cite{voigt2017eu} and the California Consumer Privacy Act (CCPA) \cite{goldman2020introduction}, which impose strict constraints on the collection and processing of personal information. These regulations necessitate privacy-aware video understanding systems that strike a balance between task performance and compliance requirements. As these models are increasingly used in real-world applications, it becomes crucial to address the privacy risks associated with video representations, particularly without compromising task performance.

Despite the increasing use of advanced video representation models, many early privacy-preserving approaches address privacy risks by directly modifying the input video data, rather than the learned representations. A common strategy is to reduce the spatial resolution of video frames through downsampling, which limits the visibility of identifiable details \cite{dai2015towards, liu2020indoor, ryoo2017privacy}. Another approach involves detection-based anonymization, in which pretrained detectors locate privacy-sensitive areas in the video. These areas are then obscured using techniques such as blurring \cite{zhang2021multi} or replaced with synthetic content \cite{ren2018learning}. While these methods can reduce visual identifiability, they often fail to anonymize motion patterns in video data, which can still lead to identity leakage. At the same time, aggressive spatial modifications disrupt the spatiotemporal structure required for effective video understanding. As a result, such input-level anonymization strategies both provide limited privacy protection and degrade the performance of downstream recognition tasks. 

More recent work has shifted toward learning-based privacy-preserving methods, where anonymizers are trained in an adversarial manner to suppress private attributes while maintaining task performance \cite{dave2022spact, fioresi2023ted, wu2020privacy, aslam2025balancing}. These approaches have shown improved results compared to input-level anonymization, particularly in removing appearance-based privacy cues. However, many of them still remain vulnerable to privacy leakage through motion patterns, which can encode identity-related information over time. In contrast, STPrivacy \cite{li2023stprivacy} proposes a video anonymization framework that jointly considers appearance and temporal information. It represents videos as spatiotemporal tubelets and applies token sparsification together with embedding-space anonymization under an adversarial objective. While effective, the token selection process in STPrivacy does not explicitly disentangle utility and privacy signals. As a result, the same tubelets may simultaneously contribute to both action recognition and privacy inference, making it difficult to precisely control the privacy–utility trade-off. Motivated by this limitation, we propose an attention-driven spatiotemporal video anonymization framework that explicitly separates utility and privacy objectives. Our method introduces dedicated classification tokens for action recognition and privacy prediction, and leverages their attention responses to assign each tubelet a utility-privacy score. We then retain the top-$k$ tubelets with the highest scores, thereby pruning tubelets dominated by privacy cues while preserving those critical for action recognition. This allows us to achieve competitive action recognition performance while removing data prone to leaking private information.

The contributions of the proposed work are summarized as follows:

\begin{enumerate}
    \item We introduce two task-specific classification tokens, \(\texttt{[act\_CLS]}\) and \(\texttt{[priv\_CLS]}\), with mutual attention masking. Each token attends to all spatiotemporal tubelets while remaining isolated from the others.

    \item We propose an attention-driven tubelet scoring function,
    \(s_i = \alpha_i^{\text{act}} - \lambda_{\text{priv}} \cdot \alpha_i^{\text{priv}}\),
    where \(\alpha_i^{\text{act}}\) and \(\alpha_i^{\text{priv}}\) are the attention weights assigned to tubelet \(i\) by the action and privacy tokens, respectively, and $s_i$ is the resulting utility--privacy attention score used for top-$k$ tubelet selection.

    \item We conduct extensive experiments on publicly available datasets to validate the effectiveness of the proposed framework under both in-dataset and cross-dataset evaluation settings.
\end{enumerate}


\section{Related Work}
\label{sec:related_work}

The developments in privacy-preserving vision systems can be broadly categorized into two main approaches: 1) Traditional methods and 2) Learning-based methods.

Traditional privacy-preserving methods include techniques such as blurring, pixelation, downsampling, masking, and obfuscation \cite{chou2018privacy,liu2020indoor,butler2015privacy,jaichuen2023blur}. These approaches aim to reduce privacy attributes while retaining enough visual context for downstream tasks. However, these methods have two significant limitations. First, they require domain knowledge to accurately identify areas of interest. Second, they do not operate as end-to-end processes; instead, they consist of two distinct steps: detecting or segmenting private objects and then modifying or removing them \cite{ren2018learning}. Additionally, these methods often lead to significant degradation of image quality, which can hinder downstream tasks such as expression analysis and attribute prediction \cite{newton2005preserving}. Previous studies have also shown that these anonymization techniques can be partially reversible, raising concerns about their effectiveness in adversarial settings and the risk of privacy leakage \cite{mcpherson2016defeating}. Therefore, these approaches may not be sufficient for situations that demand both strong privacy protection and high utility in downstream tasks.

Learning-based approaches employ an end-to-end training pipeline to optimize anonymization models, allowing for a trade-off between the downstream task performance and the privacy preservation. Wu \etal \cite{wu2020privacy} introduced an adversarial training framework that uses privacy labels to train the anonymizer. Building on this idea, MaSS \cite{chen2022mass} proposed a similar adversarial framework with a composite loss that selectively retains certain attributes instead of completely obscuring them. Further, some research has employed diffusion-based frameworks for face anonymization \cite{kung2025face, piano2024latent}. Dave \etal \cite{dave2022spact} introduced a self-supervised privacy-preserving framework that eliminates the need for labeled private attributes. Similarly, the self-supervised adversarial anonymization framework proposed for anomaly detection does not rely on privacy labels and employs an NT-Xent contrastive loss in the budget branch to minimize spatial privacy leakage \cite{fioresi2023ted}. Aslam \etal \cite{aslam2025balancing} propose a penalty-driven self-supervised adversarial training framework that discourages the anonymizer from encoding private attributes during training. Moving beyond image-level anonymization methods, Li \etal introduced STPrivacy \cite{li2023stprivacy}, a video anonymization framework that removes privacy-sensitive tubelets from input videos. The framework trains a video transformer in an adversarial manner to suppress privacy-related attributes. However, the token selection process in STPrivacy is not explicitly guided by a disentangled signal for utility versus privacy. As a result, controllability is limited when the same tubelets contribute to both action recognition and privacy inference. In contrast, our approach introduces two distinct classification tokens for action recognition and privacy attribute prediction, respectively. By leveraging their attention responses, we compute tubelet importance based on the relative contribution to utility and privacy, enabling more precise and controllable privacy-aware token pruning.


\begin{figure*}[h]
    \centering
    \includegraphics[width=\linewidth]{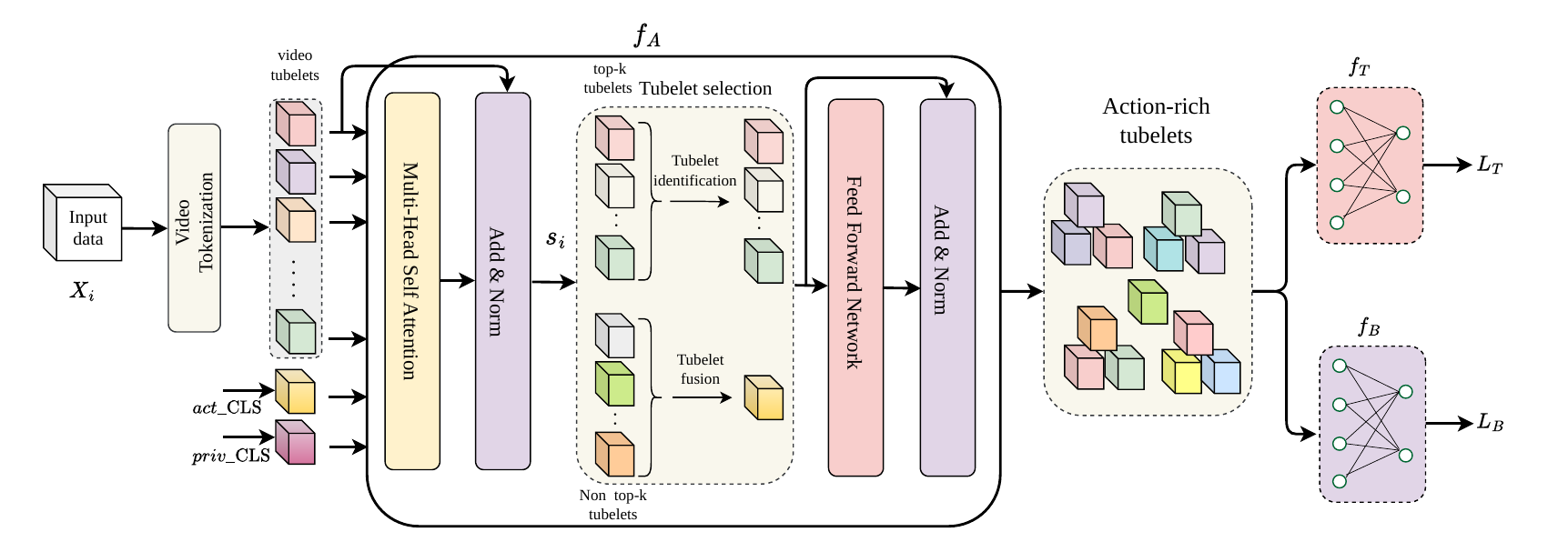}
    \caption{Overview of our proposed video anonymization framework. A ViT-based anonymizer $f_A$ employs two independent CLS tokens, \texttt{act\_CLS} and \texttt{priv\_CLS}, to attend all tubelet sequence. Each tubelet $i$ is assigned a utility-privacy score $s_i=\alpha_i^{\text{act}} -\lambda_{\text{priv}} \cdot \alpha_i^{\text{priv}}$, where $\alpha_i^{\text{act}}$ and $\alpha_i^{\text{priv}}$ denote the CLS-to-tubelet attention weights. Tubelets are then ranked by ${s_i}$ and the top-$k$ tubelets are retained to suppress privacy-sensitive content while preserving action-relevant cues, while the remaining low-score tubelets are compressed via a token fusion module. The anonymized video is then fed to a utility branch $f_T$ and a budget model $f_B$ under action and privacy objectives.}
    \label{architecture}
\end{figure*}

\section{Method}
\label{sec:methodology}

The objective of our attention-driven spatiotemporal anonymization framework is to suppress spatiotemporal video patterns that leak privacy while preserving action-discriminative cues in achieving competitive action recognition performance. To this end, we introduce two task-specific classification tokens, $\texttt{[act\_CLS]}$ and $\texttt{[priv\_CLS]}$, within a ViT-based anonymizer. During training, we leverage their attention maps to compute a token-level utility-privacy score for tubelet selection and prune tubelets dominated by privacy cues while retaining those most relevant to action recognition. This design maintains task utility under stronger privacy constraints. The overall architecture of the proposed framework is shown in Figure.~\ref{architecture}.

\subsection{Problem Statement}

We formulate our problem similar to the existing privacy-preserving action recognition frameworks \cite{dave2022spact, wu2020privacy, aslam2025balancing}, but with a different optimization objective. Given an input video $X$, we consider two predictive functions: an action recognition model $f_T$, which defines the utility task, and a private‑attribute classifier $f_B$, which defines the budget task. Our goal is to learn an anonymization function $f_A$ that minimizes privacy leakage, ensuring the minimal degradation in the utility task. To this end, we optimize $f_A$ toward an optimal anonymizer, $f_A^*$, which keeps features for utility task while removing features for budget task. To achieve this, we have introduced two task-specific classification tokens \texttt{[act\_CLS]} and \texttt{[priv\_CLS]} in a ViT-based anonymizer. These tokens explicitly disentangle action-relevant and privacy-relevant spatiotemporal representations during the training and guide the anonymizer to remove the features associated with the privacy budget task while minimally affecting the representations related to utility task. This can be formulated by the following two criteria:

\textit{\textbf{Criterion 1}}: The goal of the learned anonymizer, $f_A^*$, is to preserve utility-relevant features such that the performance of the action recognition model $f_T'$ on anonymized input remains close to that on the raw video.
\begin{equation}
    L_T(f_T'(f_A^*(X), Y_T)) \approx L_T(f_T'(X), Y_T),
    \label{c1}
\end{equation}
Here $L_T$ is the loss function of the utility task, $f_T'$ is the utility task model, and $Y_T$ is the label associated to the utility task.

\textit{\textbf{Criterion 2}}: The learned anonymizer, $f_A^*$, effectively reduces privacy leakage, resulting in a significant decline in budget performance compared to the raw data. Mathematically, this can be expressed as:
\begin{equation}
    L_B\big(f_B'(f_A^*(X)),\ Y_B\big) \gg L_B\big(f_B'(X),\ Y_B\big),
    \label{c2}
\end{equation}
where $L_B$ is the loss function of the budget task, $f_B'$ is the privacy budget model, and $Y_B$ is private attribute labels.

In practice, satisfying Criterion~\ref{c1} is challenging, as the anonymizer lacks explicit knowledge of which features support the utility task and which reveal private attributes. Since these features are entangled in the spatiotemporal domain, effective separation is essential. To address this, we design a ViT-based anonymizer with two distinct CLS tokens that disentangle utility and privacy-relevant features, enabling targeted anonymization guided by the proposed criteria.

\subsection{Proposed Approach}

\subsubsection{Video Tokenization}

Let ${(X, Y_T, Y_B)}$ denote the training dataset, where $X \in \mathbb{R}^{T \times C \times H \times W}$ is a video with temporal length $T$, channel dimension $C$, height $H$, and width $W$. The label $Y_T \in \{0,1\}^A$ is a one-hot vector over $A$ action classes representing the utility task, while $Y_B \in \{0,1\}^P$ denotes the binary privacy labels for $P$ private attributes. The $i^{\text{th}}$ entry of $Y_B$ indicates whether the corresponding private attribute is present in video $X$. To enable Transformer-based processing, we tokenize each input video into a sequence of non-overlapping spatiotemporal tubelets, where each tubelet is of size $\delta_T \times \delta_H \times \delta_W \times C$, and is embedded into a $D$-dimensional token using 3D convolutions. This results in a token sequence $x \in \mathbb{R}^{L \times N \times D}$, where $L = T / \delta_T$ is the number of temporal segments, and $N = (H / \delta_H) \cdot (W / \delta_W)$ is the number of spatial tokens per segment. The spatial and temporal dimensions are flattened before feeding the tokens into the transformer.

To enable disentangled feature learning for utility and privacy, we prepend two task-specific classification tokens, \(\texttt{[act\_CLS]}\) and \(\texttt{[priv\_CLS]}\), to the token sequence. These tokens attend to all tubelet tokens via self-attention, but do not attend to each other. They are optimized for the utility and privacy objectives, respectively. The anonymizer leverages their attention responses to identify and suppress tubelets that are privacy-relevant, while preserving those essential for the utility task.

\subsubsection{Tubelet Identification for Selective Anonymization}

To achieve the stated objectives in Criterion~\ref{c1} and ~\ref{c2}, we propose an attention-driven token selection mechanism which is based on task-specific classification tokens. These tokens are trained for specific objectives: \texttt{[act\_CLS]} aggregates features relevant to the utility task, while \texttt{[priv\_CLS]} focuses on features associated with the privacy budget task. Each token receives task-specific supervision through its corresponding classification head. 

To enforce task disentanglement, we compute the attention scores for the utility task such that \texttt{[act\_CLS]} attends to all tubelet tokens except \texttt{[priv\_CLS]}. Conversely, the attention scores for the privacy budget task are computed by allowing \texttt{[priv\_CLS]} to attend to all tubelet tokens except \texttt{[act\_CLS]}. This design enables each CLS token to compute an independent attention distribution over the shared input while remaining isolated from the influence of the other. This behavior is implemented via a binary attention mask applied before the softmax operation in the self-attention mechanism, ensuring that no cross-attention occurs between the two CLS tokens. At each Transformer layer, we compute multi-head self-attention maps from both CLS tokens to the tubelet tokens. For attention head \(h\), let \(attn^{(h)}_{act} \in \mathbb{R}^{L \times N}\) and \(attn^{(h)}_{priv} \in \mathbb{R}^{L \times N}\) denote the softmax-normalized attention weights from \texttt{[act\_CLS]} and \texttt{[priv\_CLS]} to all tubelet tokens, respectively. We then average these attention weights across all heads to obtain global, task-specific attention weights:

\begin{equation}
    \alpha_{i}^{\text{act}} = \frac{1}{H}\sum_{h=1}^{H} attn_{\text{act}}^{(h)}(i),
    \qquad
    \alpha_{i}^{\text{priv}} = \frac{1}{H}\sum_{h=1}^{H} attn_{\text{priv}}^{(h)}(i).
\end{equation}

Using these attention weights, we compute a privacy-aware tubelet score:

\begin{equation}
    s_i = \alpha_{i}^{\text{act}} - \lambda_{\text{priv}}\,\alpha_{i}^{\text{priv}},
\end{equation}

where $\lambda_{\text{priv}}$ is a hyperparameter controlling the strength of privacy suppression and set as 0.5 to enforce a balanced utility–privacy trade-off in our experimental set up. Tubelets with higher scores are more relevant for the utility objective while being less indicative of private attributes.

Based on $\{s_i\}_{i=1}^{L \times N}$, we perform top-$k$ tubelet selection by retaining the indices of the $k$ highest-scoring tubelets as:

\begin{equation}
    \mathcal{I}_k = \operatorname{TopK}\big(\{s_i\}_{i=1}^{L \times N},\,k\big),
\end{equation}

This selective tubelet pruning preserves action discriminative information while suppressing tubelets most related with privacy attributes. Unlike predefined masking or region-based removal, our approach performs dynamic, input-adaptive suppression through end-to-end optimization guided by disentangled task attention.

\subsubsection{Tubelet Fusion}

While top-$k$ tubelet selection removes tubelets that are less favorable under the utility-privacy criterion, fully discarding the remaining tubelets can eliminate subtle contextual cues that still support action recognition, particularly in early training. To mitigate this, we adopt an inattentive tubelet fusion strategy inspired by~\cite{liang2022not}, due to its strong performance on image classification datasets \cite{haurum2023iccv}. Let $\mathcal{T}_{\text{keep}}$ denote the set of top-$k$ selected tubelets and $\mathcal{T}_{\text{drop}}$ the remaining tubelets. Instead of discarding $\mathcal{T}_{\text{drop}}$, we compress them into a single residual tubelet via an action attention weighted sum:

\begin{equation}
    t_{\text{fuse}} = \sum_{i \in \mathcal{T}{\text{drop}}} \alpha_i^{\text{act}} \cdot x_i
\end{equation}

where $x_i$ is the feature of tubelet $i$ and $\alpha_i^{\text{act}}$ is its normalized attention weight from \texttt{[act\_CLS]}. We then append $t_{\text{fuse}}$ to $\mathcal{T}_{\text{keep}}$ and forward the resulting tubelet sequence to the next Transformer layer. This fusion provides a residual pathway that retains compressed contextual information from pruned regions, improving representational coverage while maintaining the efficiency gains of tubelet pruning. Algorithm~\ref{algorithm} outlines the overall pipeline of the proposed method.


\begin{algorithm}[h]
\caption{Tubelet selection and training pipeline}
\label{algorithm}
\KwIn{$(X,Y_T,Y_B)$, keep rate $r$, pruning layers $\mathcal{P}$}
\KwOut{$\theta_A^{*}, \theta'_T, \theta'_B$}

\textbf{1. Video tokenization:}\ \\
$X \rightarrow \{x_i\}_{i=1}^{M}$\
$Z \leftarrow [\texttt{[act\_CLS]},\texttt{[priv\_CLS]},x_1,\dots,x_M]$\

\textbf{2. Selective pruning:}\ \\
\For{$\ell \leftarrow 1$ \KwTo $L$}{
$Z \leftarrow \text{Block}_\ell(Z)$\ \\
\If{$\ell \in \mathcal{P}$}{
Compute $\alpha^{\text{act}},\alpha^{\text{priv}}$\ \\
$s_i \leftarrow \alpha_i^{\text{act}} - \lambda_{\text{priv}}\alpha_i^{\text{priv}}$\ \\
        $k_\ell \leftarrow \left\lceil r\,M_{\ell-1}\right\rceil$ ,
        \quad 
        $\mathcal{I}_{k_\ell} \leftarrow \operatorname{TopK}(\{s_i\}, k_\ell)$\ \\
$Z \leftarrow [\texttt{[act\_CLS]},\texttt{[priv\_CLS]},\{x_i\}_{i\in\mathcal{I}_{k_\ell}}]$\ \\
}
}

\textbf{3. Adversarial training:}\ \\
\For{$e \leftarrow 1$ \KwTo $e_{\text{anon}}$}{
$\hat{y}_T \leftarrow \text{head}_{\text{act}}(Z[\texttt{[act\_CLS]}])$\ \\
$\hat{y}_B \leftarrow \text{head}_{\text{priv}}(Z[\texttt{[priv\_CLS]}])$\ \\
Update $\theta_A,\theta_T,\theta_B$ \
}
Freeze $\theta_A^{*}\leftarrow\theta_A$ \ \\

\textbf{4. Utility evaluation:}\ \\
Init $f'_T(\theta'_T)$; \For{$e \leftarrow 1$ \KwTo $e_{\text{act}}$}{
$\hat{y}'_T \leftarrow f'_T(f_A^{*}(X))$ \
}

\textbf{5. Privacy evalutaion:}\ \\
Init $f'_B(\theta'_B)$; \For{$e \leftarrow 1$ \KwTo $e_{\text{priv}}$}{
$\hat{y}'_B \leftarrow f'_B(f_A^{*}(X))$ \
}
\end{algorithm}

\subsection{Optimization and Objectives}

In the process of performing the privacy-aware training objectives, we follow three steps.

\subsubsection{Initialization}

We initialize the anonymizer $f_A$ using a ViT backbone that has been pretrained on ImageNet. This approach helps accelerate convergence and enhances feature learning. We extend the spatial patch embeddings into spatiotemporal tubelets by repeating kernels along the temporal axis. Positional embeddings are interpolated and replicated to match the tokenized video input.

\subsubsection{Adversarial Training}

To enforce privacy suppression while preserving utility, we adopt an adversarial training strategy using a Gradient Reversal Layer \cite{wu2020privacy, dave2022spact, aslam2025balancing} for the privacy budget task. The anonymizer is optimized by the following loss function:
\begin{equation}  
    \mathcal{L}_{\text{adv}} = L_T(f_T(f_A(X), Y_T)) + L_B(f_B(f_A(X), Y_B))
\end{equation}  
where $L_T$ is the cross-entropy loss for action classification, $L_B$ is the binary cross-entropy loss for private attribute.

\subsubsection{Evaluation}

To evaluate the effectiveness of our proposed anonymizer model, we first freeze the parameters of the trained anonymizer, $f_A^*$, and use it to convert the raw video data. Next, we train an action recognizer, $f_T^{\prime}$, and a privacy attribute classifier, $f_B^{\prime}$, using the anonymized training set. We then assess the performance of both models on their respective test sets, measuring how effectively the anonymized videos retain action-relevant information while minimizing the inclusion of privacy-sensitive content.


\section{Experiments}
\label{sec:experiments}


\begin{table*}[h]
\centering
\resizebox{\textwidth}{!}{
\begin{tabular}{c|ccc|ccc|ccc}
\hline
\multirow{3}{*}{Method} 
& \multicolumn{3}{c|}{\textbf{VPUCF}}  
& \multicolumn{3}{c|}{\textbf{VPHMDB}}  
& \multicolumn{3}{c}{\textbf{PAHMDB}}                                 
\\ 
& Action & \multicolumn{2}{c|}{Privacy} 
& Action & \multicolumn{2}{c|}{Privacy} 
& Action & \multicolumn{2}{c}{Privacy} 
\\ 
& Top-1 ($\uparrow\%$) & cMAP ($\downarrow\%$) & F1 ($\downarrow$) 
& Top-1 ($\uparrow\%$) & cMAP ($\downarrow\%$) & F1 ($\downarrow$) 
& Top-1 ($\uparrow\%$) & cMAP ($\downarrow\%$) & F1 ($\downarrow$) 
\\
\hline
Raw data              & 84.88 & 76.62 & 0.684 & 81.05 & 76.62 & 0.684 & 65.24 & 70.20 & 0.396   
\\
Downsample-2$\times$  & 54.11 & 57.31 & 0.512 & 40.80 & 71.35 & 0.601 & 36.10 & 61.20 &  0.111
\\
Downsample-4$\times$  & 39.65 & 52.61 & 0.493 & 31.32 & 69.79 & 0.594 & 25.80 & 41.40 &  0.081
\\
Blackening            & 53.13 & 56.39 & 0.457 & 38.27 & 74.06 & 0.649 & 34.20 & 63.80 &  0.386
\\
StrongBlur            & 55.59 & 55.94 & 0.456 & 40.91 & 74.33 & 0.655 & 36.40 & 64.40 &  0.243
\\
\hline
VITA \cite{wu2020privacy}             & 62.10 (\textcolor{blue}{$\downarrow$22.78}) & 55.32 (\textcolor{blue}{$\downarrow$21.3}) & 0.461 (\textcolor{blue}{$\downarrow$0.223}) & 48.11 (\textcolor{blue}{$\downarrow$32.94}) & 73.89 (\textcolor{blue}{$\downarrow$2.73}) & 0.638 (\textcolor{blue}{$\downarrow$0.046}) & 42.30 (\textcolor{blue}{$\downarrow$22.94}) & \textbf{62.30} (\textcolor{blue}{$\downarrow$7.9}) & 0.194 (\textcolor{blue}{$\downarrow$0.202}) 
\\
SPAct \cite{dave2022spact}            & 62.03 (\textcolor{blue}{$\downarrow$22.85}) & 57.43 (\textcolor{blue}{$\downarrow$19.19}) & 0.473 (\textcolor{blue}{$\downarrow$0.211}) & --- & --- & --- & 43.10 (\textcolor{blue}{$\downarrow$22.14}) & 62.70 (\textcolor{blue}{$\downarrow$7.5}) &  \textbf{0.176} (\textcolor{blue}{$\downarrow$0.220})
\\
Balancing; B=0.3 \cite{aslam2025balancing}            & --- & --- & --- & 74.61 (\textcolor{blue}{$\downarrow$6.44}) & \textbf{70.01} (\textcolor{blue}{$\downarrow$6.61}) & \textbf{0.514} (\textcolor{blue}{$\downarrow$0.170}) & 60.58 (\textcolor{blue}{$\downarrow$4.66}) & 65.67 (\textcolor{blue}{$\downarrow$4.53}) & 0.253 (\textcolor{blue}{$\downarrow$0.143})
\\
STPrivacy \cite{li2023stprivacy}      & \textbf{82.55} (\textcolor{blue}{$\downarrow$2.33}) & 73.79 (\textcolor{blue}{$\downarrow$2.83}) & 0.634 (\textcolor{blue}{$\downarrow$0.05}) & 50.73 (\textcolor{blue}{$\downarrow$30.32}) & 72.48 (\textcolor{blue}{$\downarrow$4.14}) & 0.613 (\textcolor{blue}{$\downarrow$0.071}) & 50.61 (\textcolor{blue}{$\downarrow$14.63}) & 68.76 (\textcolor{blue}{$\downarrow$1.44}) & 0.323 (\textcolor{blue}{$\downarrow$0.073})
\\
\hline
\textbf{Ours}            & 80.96 (\textcolor{blue}{$\downarrow$3.92}) & \textbf{60.30} (\textcolor{blue}{$\downarrow$16.32}) & \textbf{0.531} (\textcolor{blue}{$\downarrow$0.153}) & \textbf{79.59} (\textcolor{blue}{$\downarrow$1.46}) & 70.32 (\textcolor{blue}{$\downarrow$6.3}) & 0.562 (\textcolor{blue}{$\downarrow$0.122}) & \textbf{62.58} (\textcolor{blue}{$\downarrow$2.66}) & 65.92 (\textcolor{blue}{$\downarrow$4.28}) &  0.296 (\textcolor{blue}{$\downarrow$0.100})
\\
\hline
\end{tabular}
}
\caption{Comparison of different privacy-preserving methods on \textbf{known action} and \textbf{known private} attributes. \textcolor{blue}{$\downarrow$\%} is the relative drop from the raw data, and – indicates the model does not perform the experiment on the dataset. High performance in action and low performance in privacy are considered as better. Best results are in bold.}
\label{tab:same_results}
\end{table*}



\begin{table*}[h]
\centering
\resizebox{13cm}{!}{
\begin{tabular}{c|ccc|ccc}
\hline
\multirow{3}{*}{Method} 
& \multicolumn{3}{c|}{\textbf{VPUCF $\rightarrow$ VPHMDB}} 
& \multicolumn{3}{c}{\textbf{VPHMDB $\rightarrow$ VPUCF}} \\ 
& Action & \multicolumn{2}{c|}{Privacy} 
& Action & \multicolumn{2}{c}{Privacy} \\ 
& Top-1 ($\uparrow$\%) & cMAP ($\downarrow$\%) & F1 ($\downarrow$) 
& Top-1 ($\uparrow$\%) & cMAP ($\downarrow$ \%) & F1 ($\downarrow$)
\\ 
\hline
Raw data     & 52.31 & 75.58 & 0.673 & 92.23 & 76.62 & 0.701 
\\ 
VITA \cite{wu2020privacy}       & --- & --- & --- & 77.48 (\textcolor{blue}{$\downarrow$14.75}) & 76.02 (\textcolor{blue}{$\downarrow$0.60}) & 0.669 (\textcolor{blue}{$\downarrow$0.032})
\\ 
SPAct \cite{dave2022spact}       & --- & --- & --- & 78.13 (\textcolor{blue}{$\downarrow$14.10}) & 75.98 (\textcolor{blue}{$\downarrow$0.64}) & 0.661 (\textcolor{blue}{$\downarrow$0.040})
\\ 
STPrivacy \cite{li2023stprivacy}  & 49.56 (\textcolor{blue}{$\downarrow$2.76}) & 71.25 (\textcolor{blue}{$\downarrow$4.33}) & 0.595 (\textcolor{blue}{$\downarrow$0.078}) & \textbf{81.04} (\textcolor{blue}{$\downarrow$11.19}) & 74.60 (\textcolor{blue}{$\downarrow$2.22}) & 0.645 (\textcolor{blue}{$\downarrow$0.056})
\\ 
\hline
\textbf{Ours} & \textbf{50.69 }(\textcolor{blue}{$\downarrow$1.62}) & \textbf{59.30} (\textcolor{blue}{$\downarrow$16.28}) & \textbf{0.554} (\textcolor{blue}{$\downarrow$0.119}) & 79.83 (\textcolor{blue}{$\downarrow$12.40}) & \textbf{69.81} (\textcolor{blue}{$\downarrow$6.81}) & \textbf{0.583} (\textcolor{blue}{$\downarrow$0.118}) 
\\ 
\hline
\end{tabular}
}
\caption{Comparison of different privacy-preserving methods on \textbf{novel actions} and \textbf{novel private} attributes datasets. \textcolor{blue}{$\downarrow$\%} represents the relative drop from raw data, and – indicates the model does not perform the experiment on the dataset. High performance in action and low performance of privacy is considered as better. Best results are in bold.}
\label{tab:novel_dataset}
\end{table*}


\subsection{Datasets:}
\textbf{VPUCF} and \textbf{VPHMDB} \cite{li2023stprivacy} are large-scale video datasets annotated for both action recognition and privacy attributes. VPHMDB is derived from the HMDB51  \cite{kuehne2011hmdb} dataset, consisting of 6,849 videos spanning 51 action categories. Likewise, VPUCF is based on UCF101 \cite{soomro2012ucf101} and includes 13,320 videos across 101 action classes. Both datasets provide video-level annotations for five privacy-related attributes: \textit{face}, \textit{skin color}, \textit{gender}, \textit{nudity}, and \textit{familiar relationships}, making them ideal for assessing privacy-preserving video analysis methods.

\textbf{PAHMDB} \cite{wu2020privacy} is a privacy-annotated subset of the HMDB51 dataset \cite{kuehne2011hmdb}, consisting of 515 videos across 51 action categories. Each video is not only labeled with action categories but also annotated at the frame level for five privacy-related attributes: \textit{skin color}, \textit{face}, \textit{gender}, \textit{nudity}, and \textit{relationship}. This detailed annotation makes PAHMDB a valuable resource for researching privacy preservation in video understanding tasks.

\subsection{Evaluation metrics}

The performance of action recognition is evaluated using the standard Top-1 accuracy metric. Privacy attribute recognition is defined as a multi-label binary classification task, and it is measured by using two metrics: class-wise mean Average Precision (cMAP) and class-wise F1 score. Both Top-1 and cMAP are reported in percentages, while F1 scores are presented as decimal values.

\subsection{Implementation details:}

The anonymizer $f_A$ utilizes a ViT architecture, operating in an encoder-style manner to process spatiotemporal tubelet tokens and perform task-guided token pruning and anonymization \cite{liang2022not}. Each input video of size $16 \times 224 \times 224 \times 3$ is divided into non-overlapping tubelets of size $2 \times 16 \times 16 \times 3$ across 16 frames. Each tubelet is then embedded into a 768-dimensional token. The Transformer encoder comprises 12 layers. To enhance regularization and convergence, we apply a dropout rate of $0.2$, an attention dropout rate of $0.1$, and stochastic depth with a drop path rate of $0.2$. Tubelet pruning occurs at layers 3, 6, and 9, at a keep rate 0.9. The anonymizer $f_A$ is trained for 80 epochs using the Adam optimizer \cite{kingma2014adam} with a batch size of $8$.

For evaluation, we use an I3D ResNet-50 \cite{carreira2017quo} model as the utility action recognizer $f_T$, and a standard ResNet-50 \cite{he2016deep} as the privacy attribute classifier $f_B$. The utility classifier $f_T$ is initialized with weights pretrained on Kinetics-400, and the privacy classifier $f_B$ is initialized with ImageNet weights. For evaluation, $f_T$ and $f_B$ is trained for 50 epochs with a batch size of $16$, beginning with an initial learning rate of $1\text{e}{-3}$, utilizing linear warm-up and step decay strategies. All experiments are conducted using the PyTorch framework on NVIDIA TESLA V100 GPUs with a fixed random seed of $42$.

\subsection{Comparison with state-of-the-art methods:}

To ensure fair comparisons, we evaluate all state-of-the-art methods using the same action and video privacy recognizers within our framework. The Raw baseline is established using the original videos. Additionally, we include reference baselines such as Downsample-2$\times$ and Downsample-4$\times$, as well as blackening and strong blur. However, we exclude these from the main comparisons because they significantly affect action performance.

\subsubsection{Comparison for Known actions and Private Attributes}

In this experimental setup, the anonymizer $f_A$ is trained using the dataset $(X, Y_T, Y_B)$. Here, $X$ represents the training videos, $Y_T$ consists of the action labels, and $Y_B$ contains the privacy labels. The goal of the anonymizer is to retain information relevant to the actions while anonymizing cues that associated with privacy. To evaluate its performance, we utilized a new action classifier $f_T'$ on the outputs of the trained anonymizer $f_A^*(X_{\text{action}})$ with the action labels $Y_T$. Additionally, we measured privacy leakage by training a new privacy classifier $f_B'$ using the outputs from $f_A^*(X_{\text{privacy}})$  along with the privacy labels $Y_B$.

Table~\ref{tab:same_results} compares our method with state-of-the-art privacy-preserving baselines under the same evaluation protocol. On VPUCF dataset, STPrivacy~\cite{li2023stprivacy} attains the highest action performance with $82.55\%$ Top-1, but it only slightly reduces privacy leakage ($73.79\%$ cMAP, $0.634$ F1). In contrast, our method suppresses privacy cues more effectively, reducing privacy scores to $60.30\%$ cMAP and $0.531$ F1, while maintaining competitive action accuracy of $80.96\%$ Top-1. Relative to STPrivacy, our method yields a larger privacy reduction of $13.49 \%$ cMAP and $0.103$ F1, with a small additional utility drop of $1.59\%$ Top-1. 

On VPHMDB dataset, our approach provides the best overall balance across the compared methods. We preserve action performance close to the raw baseline ($79.59\%$ Top-1, corresponding to a $1.46\%$ drop) and privacy leakage among the evaluated privacy preserving baseline methods ($70.32\%$ cMAP and $0.562$ F1). Balancing \cite{aslam2025balancing} achieves a cMAP of 70.01\% and an F1 score of 0.514, which is very close to our privacy metrics, but with a lower action accuracy of 74.61\%. Also, STPrivacy \cite{li2023stprivacy} exhibits a substantial utility degradation on VPHMDB ($50.73\%$ Top-1) while still leaving relatively high privacy leakage ($72.48\%$ cMAP, $0.613$ F1), whereas our method avoids this utility collapse and improves privacy suppression simultaneously. Earlier baselines such as VITA~\cite{wu2020privacy} and SPAct~\cite{dave2022spact} show a weaker trade-off, typically incurring large drops in action accuracy with limited privacy reduction. Moreover, these methods are primarily designed for image-level anonymization, while our approach operates on videos and explicitly targets spatiotemporal privacy cues. 

On PAHMDB dataset, our method attains the highest action accuracy among all evaluated methods (62.58\% Top-1) while reducing privacy leakage to 65.92\% of cMAP and 0.296 of F1 score. While some baseline methods show stronger privacy suppression, such as VITA \cite{wu2020privacy}, which has a cMAP of 62.30\% and an F1 score of 0.194, and SPAct \cite{dave2022spact}, which reports an F1 score of 0.176, these methods come with significantly lower utility, with Top-1 accuracies of only 42.30\% and 43.10\%, respectively. This emphasizes the advantage of our approach, which effectively preserves action utility while still achieving meaningful reductions in privacy leakage.
Overall, the results across all three datasets demonstrate that our attention-driven tubelet pruning significantly improves the privacy-utility trade-off. It consistently maintains high action accuracy while achieving substantial reductions in privacy leakage compared to recent video anonymization baselines.

\subsubsection{Comparison for Novel Actions and Private Attributes}

In this experiment, we evaluate the trained anonymizer $f^*_A$ on an unseen dataset to assess cross-dataset generalization. Specifically, $f^*_A$ is trained on one dataset and then tested on a different dataset, ensuring that the evaluation samples are not observed during training. Table~\ref{tab:novel_dataset} compares our framework with learning-based baselines under cross-dataset transfer, covering novel actions and novel privacy attributes. For the transfer setting VPUCF $\rightarrow$ VPHMDB, our method achieves a better utility-privacy trade-off than STPrivacy~\cite{li2023stprivacy}, with higher action accuracy ($50.69\%$ vs. $49.56\%$ Top-1) and substantially lower privacy leakage ($59.30\%$ cMAP and $0.554$ F1 vs. $71.25\%$ cMAP and $0.595$ F1). For the reverse transfer VPHMDB $\rightarrow$ VPUCF, STPrivacy attains the highest action Top-1 ($81.04\%$), while our method provides stronger privacy suppression, reducing leakage to $69.81\%$ cMAP and $0.583$ F1, compared to $74.60\%$ cMAP and $0.645$ F1 for STPrivacy, with comparable action performance ($79.83\%$ Top-1). Overall, these results indicate that our anonymizer generalizes better in terms of privacy removal across datasets, while preserving competitive action utility.

\begin{figure*}[h]
    \centering
    \begin{subfigure}[t]{\linewidth}
        \centering
        \includegraphics[width=\linewidth]{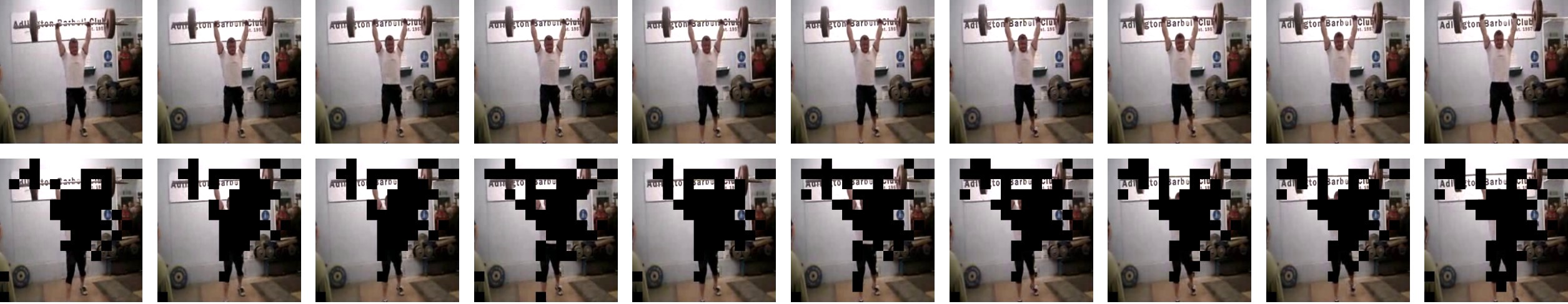}
        \caption{}
    \end{subfigure}
    \begin{subfigure}[t]{\linewidth}
        \centering
        \includegraphics[width=\linewidth]{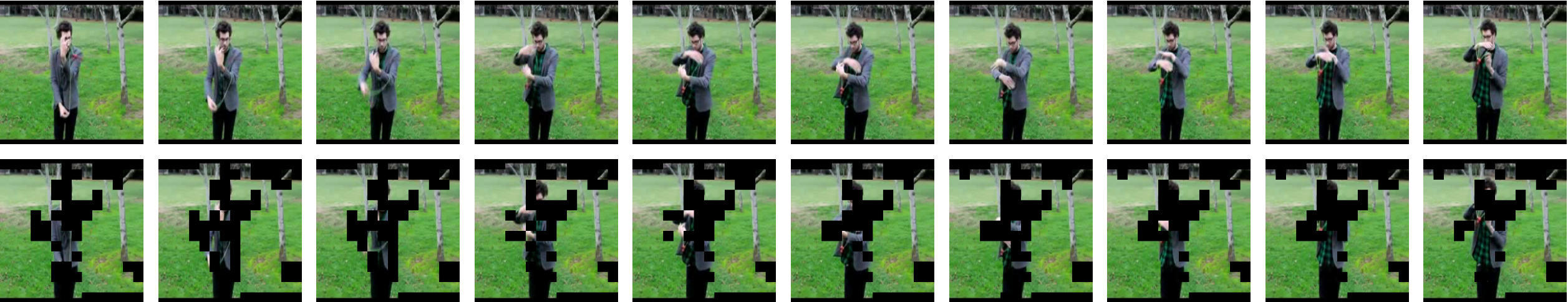}
        \caption{}
    \end{subfigure}
    \begin{subfigure}[t]{\linewidth}
        \centering
        \includegraphics[width=\linewidth]{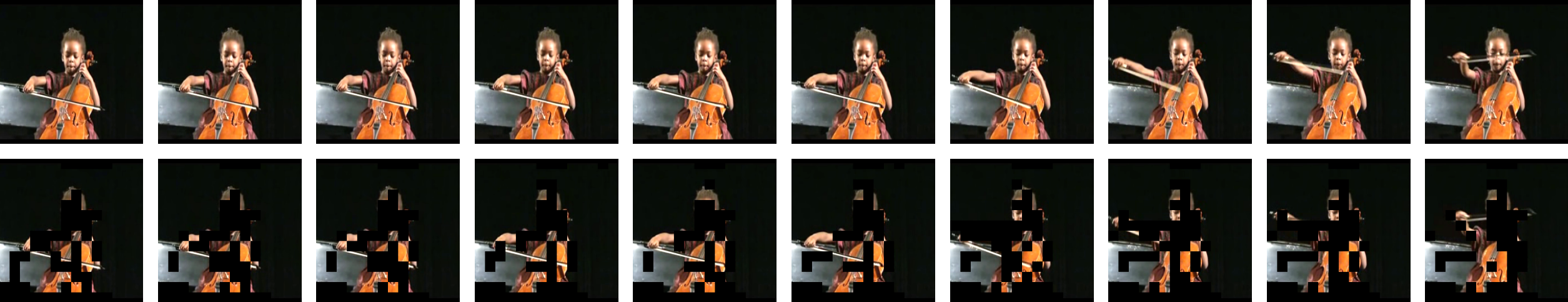}
        \caption{}
    \end{subfigure}
    \caption{Visualization of our framework on three actions: (a) weightlifting, (b) YoYo, and (c) playing violin. For each ten frame clip, the top row shows raw consecutive frames, and the bottom row shows the corresponding anonymized frames produced by our method.}
    \label{anonymized_clips}
\end{figure*}


\subsection{Qualitative Analysis}

To qualitatively validate the effectiveness of our attention-driven spatiotemporal token pruning method, Figure \ref{anonymized_clips} compares raw ten frame clips with the corresponding anonymized ten frame outputs. 

The qualitative results in Figure \ref{anonymized_clips} illustrate how the proposed attention-driven anonymization removes the privacy sensitive visual cues while preserving action relevant information. Across the three actions (weightlifting, YoYo and playing violin), the anonymized clips reduce identifiable appearance details such as facial features, skin tone, and gender, yet retain the motion structure and coarse spatial layout needed to recognize the activity (\eg, arm trajectories, object interaction, and body pose dynamics). This behavior is consistent with the design of our two CLS-token attention mechanism, which prioritizes tubelets that contribute to utility while down weighting those associated with privacy inference, and with token fusion that aggregates the remaining low-score regions without reintroducing sensitive detail. Overall, the visual comparisons support our quantitative findings by showing that privacy leakage can be reduced without eliminating the cues that define the underlying action.

\subsection{Ablation Analysis}

\subsubsection{Effects of different pruning levels}

We conduct an ablation study on the different tubelet pruning levels on VPHMDB dataset to quantify how tubelet retention affects the utility-privacy trade-off, see Table~\ref{tab:ablation_prune_level}.

Without pruning (keep rate $=1.0$, \ie, retaining all tubelets), the baseline achieves $81.05\%$ Top-1 accuracy on the utility branch, while the budget task exhibits a privacy leakage of $76.62\%$ cMAP. With mild pruning (keep rate $=0.9$, \ie, pruning $10\%$ tubelets at each pruning layer), our method preserves near baseline utility ($79.59\%$ Top-1, a $1.46\%$ absolute drop) while reducing privacy leakage to $70.32\%$ cMAP (a $6.30\%$ absolute decrease). This suggests that the tubelets removed by our scoring function are enriched with privacy-sensitive cues, whereas the retained tubelets remain largely action-discriminative for downstream recognition.

In contrast, more aggressive pruning yields diminishing privacy gains relative to the utility loss. Reducing the keep rate to $0.8$ further lowers privacy leakage to $60.32\%$ cMAP (an additional $10.00\%$ decrease compared to keep rate $=0.9$), but action accuracy drops substantially to $69.03\%$ (a $10.56\%$ decrease compared to keep rate $=0.9$). This trend becomes more pronounced at keep rate $=0.7$ (pruning $30\%$ tubelets per pruning layer), where cMAP decreases to $55.81\%$ but Top-1 accuracy falls to $60.08\%$, indicating that excessive pruning increasingly removes action-discriminative tubelets alongside privacy-sensitive ones. Overall, keep rate $=0.9$ emerges as a strong operating point, providing meaningful privacy suppression with minimal degradation in action recognition, whereas heavier pruning substantially compromises utility for comparatively smaller incremental privacy gains.


\begin{table}[h]
\centering
\begin{tabular}{@{}lll@{}}
\toprule
Keep rate $r$   & Top-1 (\%)         & cMAP (\%)          \\ 
\midrule
1.0 & 81.05         & 76.62         \\
0.9 & 79.59 ($\downarrow$1.46)  & 70.32 ($\downarrow$6.30)  \\
0.8 & 69.03 ($\downarrow$12.02) & 60.32 ($\downarrow$16.30) \\
0.7 & 60.08 ($\downarrow$20.97) & 55.81 ($\downarrow$20.81) \\ \bottomrule
\end{tabular}
\caption{Performance under different keep rates for the tubelet selection on VPHMDB dataset. $\downarrow$ denotes the absolute drop relative to the raw setting (when the keep rate $r=1.0$), \ie, without pruning.}
\label{tab:ablation_prune_level}
\end{table}



\section{Conclusion}
\label{sec:conclusion}

This paper proposes an attention-driven framework for spatiotemporal video anonymization that explicitly separates utility and privacy-relevant cues, allowing them to achieve independent objectives within a shared ViT backbone. By introducing two task-specific classification tokens, \texttt{[act\_CLS]} and \texttt{[priv\_CLS]}, our method generates disentangled attention signals that facilitate tubelet-level scoring and selective tubelet pruning. This mechanism effectively suppresses privacy-sensitive spatiotemporal cues while preserving information that is relevant for action recognition. To mitigate the loss of useful context from pruned regions, we further incorporate low score token fusion, which compresses discarded tubelets into a residual token guided by action attention weights. Extensive experiments on multiple benchmarks demonstrate that our approach achieves a stronger utility-privacy trade-off than existing baselines, reducing privacy leakage while maintaining competitive action recognition performance. Further cross-dataset evaluations show improved generalization in privacy suppression under distribution shifts. Overall, the proposed attention-driven pruning and fusion strategy offers an efficient and controllable solution for privacy-preserving video understanding, and provides a practical step toward deploying video analytics under stronger privacy constraints.


{\small
\bibliographystyle{ieee_fullname}
\bibliography{references}

@String(ICCV= {Int. Conf. Comput. Vis.})

@String(ECCV= {Eur. Conf. Comput. Vis.})

@String(ICIP = {IEEE Int. Conf. Image Process.})

@String(AAAI = {AAAI})

@String(ICCV  = {ICCV})

@String(ECCV  = {ECCV})

@String(ICIP  = {ICIP})

@article{wang2022internvideo,
  title={Internvideo: General video foundation models via generative and discriminative learning},
  author={Wang, Yi and Li, Kunchang and Li, Yizhuo and He, Yinan and Huang, Bingkun and Zhao, Zhiyu and Zhang, Hongjie and Xu, Jilan and Liu, Yi and Wang, Zun and others},
  journal={arXiv preprint arXiv:2212.03191},
  year={2022}
}

@inproceedings{wang2023videomae,
  title={Videomae v2: Scaling video masked autoencoders with dual masking},
  author={Wang, Limin and Huang, Bingkun and Zhao, Zhiyu and Tong, Zhan and He, Yinan and Wang, Yi and Wang, Yali and Qiao, Yu},
  booktitle={Proceedings of the IEEE/CVF conference on computer vision and pattern recognition},
  pages={14549--14560},
  year={2023}
}

@article{bardes2023v,
  title={V-jepa: Latent video prediction for visual representation learning},
  author={Bardes, Adrien and Garrido, Quentin and Ponce, Jean and Chen, Xinlei and Rabbat, Michael and LeCun, Yann and Assran, Mido and Ballas, Nicolas},
  year={2023}
}

@inproceedings{fioresi2023ted,
  title={Ted-spad: Temporal distinctiveness for self-supervised privacy-preservation for video anomaly detection},
  author={Fioresi, Joseph and Dave, Ishan Rajendrakumar and Shah, Mubarak},
  booktitle={Proceedings of the IEEE/CVF international conference on computer vision},
  pages={13598--13609},
  year={2023}
}

@inproceedings{dave2022spact,
  title={Spact: Self-supervised privacy preservation for action recognition},
  author={Dave, Ishan Rajendrakumar and Chen, Chen and Shah, Mubarak},
  booktitle={Proceedings of the IEEE/CVF Conference on Computer Vision and Pattern Recognition},
  pages={20164--20173},
  year={2022}
}

@inproceedings{li2023stprivacy,
  title={Stprivacy: Spatio-temporal privacy-preserving action recognition},
  author={Li, Ming and Xu, Xiangyu and Fan, Hehe and Zhou, Pan and Liu, Jun and Liu, Jia-Wei and Li, Jiahe and Keppo, Jussi and Shou, Mike Zheng and Yan, Shuicheng},
  booktitle={Proceedings of the IEEE/CVF International Conference on Computer Vision},
  pages={5106--5115},
  year={2023}
}

@article{wu2020privacy,
  title={Privacy-preserving deep action recognition: An adversarial learning framework and a new dataset},
  author={Wu, Zhenyu and Wang, Haotao and Wang, Zhaowen and Jin, Hailin and Wang, Zhangyang},
  journal={IEEE Transactions on Pattern Analysis and Machine Intelligence},
  volume={44},
  number={4},
  pages={2126--2139},
  year={2020},
  publisher={IEEE}
}

@article{voigt2017eu,
  title={The eu general data protection regulation (gdpr)},
  author={Voigt, Paul and Von dem Bussche, Axel},
  journal={A practical guide, 1st ed., Cham: Springer International Publishing},
  volume={10},
  number={3152676},
  pages={10--5555},
  year={2017},
  publisher={Springer}
}

@article{goldman2020introduction,
  title={An introduction to the california consumer privacy act (ccpa)},
  author={Goldman, Eric},
  journal={Santa Clara Univ. Legal Studies Research Paper},
  year={2020}
}

@inproceedings{dai2015towards,
  title={Towards privacy-preserving recognition of human activities},
  author={Dai, Ji and Saghafi, Behrouz and Wu, Jonathan and Konrad, Janusz and Ishwar, Prakash},
  booktitle={2015 IEEE international conference on image processing (ICIP)},
  pages={4238--4242},
  year={2015},
  organization={IEEE}
}

@inproceedings{liu2020indoor,
  title={Indoor privacy-preserving action recognition via partially coupled convolutional neural network},
  author={Liu, Jixin and Zhang, Leilei},
  booktitle={2020 International Conference on Artificial Intelligence and Computer Engineering (ICAICE)},
  pages={292--295},
  year={2020},
  organization={IEEE}
}

@inproceedings{ryoo2017privacy,
  title={Privacy-preserving human activity recognition from extreme low resolution},
  author={Ryoo, Michael and Rothrock, Brandon and Fleming, Charles and Yang, Hyun Jong},
  booktitle={Proceedings of the AAAI conference on artificial intelligence},
  volume={31},
  number={1},
  year={2017}
}

@inproceedings{ren2018learning,
  title={Learning to anonymize faces for privacy preserving action detection},
  author={Ren, Zhongzheng and Lee, Yong Jae and Ryoo, Michael S},
  booktitle={Proceedings of the european conference on computer vision (ECCV)},
  pages={620--636},
  year={2018}
}

@article{zhang2021multi,
  title={Multi-scale, class-generic, privacy-preserving video},
  author={Zhang, Zhixiang and Cilloni, Thomas and Walter, Charles and Fleming, Charles},
  journal={Electronics},
  volume={10},
  number={10},
  pages={1172},
  year={2021},
  publisher={MDPI}
}

@inproceedings{aslam2025balancing,
  title={Balancing Privacy and Action Performance: A Penalty-Driven Approach to Image Anonymization},
  author={Aslam, Nazia and Nasrollahi, Kamal},
  booktitle={Proceedings of the Computer Vision and Pattern Recognition Conference},
  pages={729--738},
  year={2025}
}

@inproceedings{butler2015privacy,
  title={The privacy-utility tradeoff for remotely teleoperated robots},
  author={Butler, Daniel J and Huang, Justin and Roesner, Franziska and Cakmak, Maya},
  booktitle={Proceedings of the tenth annual ACM/IEEE international conference on human-robot interaction},
  pages={27--34},
  year={2015}
}

@article{chou2018privacy,
  title={Privacy-preserving action recognition for smart hospitals using low-resolution depth images},
  author={Chou, Edward and Tan, Matthew and Zou, Cherry and Guo, Michelle and Haque, Albert and Milstein, Arnold and Fei-Fei, Li},
  journal={arXiv preprint arXiv:1811.09950},
  year={2018}
}

@inproceedings{jaichuen2023blur,
  title={BLUR \& TRACK: real-time face detection with immediate blurring and efficient tracking},
  author={Jaichuen, Tanakrit and Ren, Nanthaphop and Wongapinya, Pichai and Fugkeaw, Somchart},
  booktitle={2023 20th International Joint Conference on Computer Science and Software Engineering (JCSSE)},
  pages={167--172},
  year={2023},
  organization={IEEE}
}

@article{newton2005preserving,
  title={Preserving privacy by de-identifying face images},
  author={Newton, Elaine M and Sweeney, Latanya and Malin, Bradley},
  journal={IEEE transactions on Knowledge and Data Engineering},
  volume={17},
  number={2},
  pages={232--243},
  year={2005},
  publisher={IEEE}
}

@article{mcpherson2016defeating,
  title={Defeating image obfuscation with deep learning},
  author={McPherson, Richard and Shokri, Reza and Shmatikov, Vitaly},
  journal={arXiv preprint arXiv:1609.00408},
  year={2016}
}

@article{chen2022mass,
  title={Mass: Multi-attribute selective suppression},
  author={Chen, Chun-Fu and Hu, Shaohan and Shi, Zhonghao and Gulati, Prateek and Moriarty, Bill and Pistoia, Marco and Piuri, Vincenzo and Samarati, Pierangela},
  journal={arXiv preprint arXiv:2210.09904},
  year={2022}
}

@inproceedings{kung2025face,
  title={Face anonymization made simple},
  author={Kung, Han-Wei and Varanka, Tuomas and Saha, Sanjay and Sim, Terence and Sebe, Nicu},
  booktitle={2025 IEEE/CVF Winter Conference on Applications of Computer Vision (WACV)},
  pages={1040--1050},
  year={2025},
  organization={IEEE}
}

@article{piano2024latent,
  title={Latent diffusion models for attribute-preserving image anonymization},
  author={Piano, Luca and Basci, Pietro and Lamberti, Fabrizio and Morra, Lia},
  journal={arXiv preprint arXiv:2403.14790},
  year={2024}
}

@article{liang2022not,
  title={Not all patches are what you need: Expediting vision transformers via token reorganizations},
  author={Liang, Youwei and Ge, Chongjian and Tong, Zhan and Song, Yibing and Wang, Jue and Xie, Pengtao},
  journal={arXiv preprint arXiv:2202.07800},
  year={2022}
}

@inproceedings{kuehne2011hmdb,
  title={HMDB: a large video database for human motion recognition},
  author={Kuehne, Hildegard and Jhuang, Hueihan and Garrote, Est{\'\i}baliz and Poggio, Tomaso and Serre, Thomas},
  booktitle={2011 International conference on computer vision},
  pages={2556--2563},
  year={2011},
  organization={IEEE}
}

@article{soomro2012ucf101,
  title={Ucf101: A dataset of 101 human actions classes from videos in the wild},
  author={Soomro, Khurram and Zamir, Amir Roshan and Shah, Mubarak},
  journal={arXiv preprint arXiv:1212.0402},
  year={2012}
}

@inproceedings{carreira2017quo,
  title={Quo vadis, action recognition? a new model and the kinetics dataset},
  author={Carreira, Joao and Zisserman, Andrew},
  booktitle={proceedings of the IEEE Conference on Computer Vision and Pattern Recognition},
  pages={6299--6308},
  year={2017}
}

@inproceedings{he2016deep,
  title={Deep residual learning for image recognition},
  author={He, Kaiming and Zhang, Xiangyu and Ren, Shaoqing and Sun, Jian},
  booktitle={Proceedings of the IEEE conference on computer vision and pattern recognition},
  pages={770--778},
  year={2016}
}

@article{kingma2014adam,
  title={Adam: A method for stochastic optimization},
  author={Kingma, Diederik P},
  journal={arXiv preprint arXiv:1412.6980},
  year={2014}
}

@InProceedings{haurum2023iccv,
    author    = {Haurum, Joakim Bruslund and Escalera, Sergio and Taylor, Graham W. and Moeslund, Thomas B.},
    title     = {Which Tokens to Use? Investigating Token Reduction in Vision Transformers},
    booktitle = {Proceedings of the IEEE/CVF International Conference on Computer Vision (ICCV) Workshops},
    month     = {October},
    year      = {2023},
    pages     = {773-783}
}
}

\end{document}